\documentclass[letterpaper, 10 pt, conference]{ieeeconf}  

\IEEEoverridecommandlockouts                              

\usepackage{balance}
\usepackage{graphicx}
\usepackage{float}
\usepackage{multirow}
\usepackage{subfigure}
\usepackage{caption} 
\usepackage{epsfig}
\usepackage{times}
\usepackage{amsmath}
\usepackage{amssymb}
\usepackage{amsfonts}
\usepackage{mathrsfs}
\usepackage{hyperref}
\usepackage{setspace}

\title{\LARGE \bf
SCRIMP: Scalable Communication for Reinforcement- and Imitation-Learning-Based Multi-Agent Pathfinding
}

\author{Yutong Wang$^{1}$, Bairan Xiang$^{1}$, Shinan Huang$^{1}$, Guillaume Sartoretti$^{1}$$^{\dagger}$%
\thanks{$\dagger$ Corresponding author, to whom correspondence should be addressed.}
\thanks{$^{1}$Authors are with the Department of Mechanical Engineering, College of Design and Engineering, National University of Singapore, 21 Lower Kent Ridge Rd, Singapore
{\tt\small \{e0576114, e1011060, e1010775\}@u.nus.edu, mpegas@nus.edu.sg}
}}

\begin{document}
\maketitle
\thispagestyle{empty}
\pagestyle{empty}

\begin{abstract}

Trading off performance guarantees in favor of scalability, the Multi-Agent Path Finding (MAPF) community has recently started to embrace Multi-Agent Reinforcement Learning (MARL), where agents learn to collaboratively generate individual, collision-free (but often suboptimal) paths.
Scalability is usually achieved by assuming a local field of view (FOV) around the agents, helping scale to arbitrary world sizes.
However, this assumption significantly limits the amount of information available to the agents, making it difficult for them to enact the type of joint maneuvers needed in denser MAPF tasks.
In this paper, we propose SCRIMP, where agents learn individual policies from even very small (down to 3x3) FOVs, by relying on a highly-scalable global communication mechanism based on a modified transformer.
We further equip agents with a state-value-based tie-breaking strategy to further improve performance in symmetric situations, and introduce intrinsic rewards to encourage exploration while mitigating the long-term credit assignment problem.
Empirical evaluations on a set of experiments indicate that SCRIMP can achieve higher performance with improved scalability compared to other state-of-the-art learning-based MAPF planners with larger FOVs, and even yields similar performance as a classical centralized planner in many cases.
Ablation studies further validate the effectiveness of our proposed techniques.
Finally, we show that our trained model can be directly implemented on real robots for online MAPF through high-fidelity simulations in gazebo.

\end{abstract}

\section{Introduction}

Multi-Agent Path Finding (MAPF) is an NP-hard problem (even for suboptimal paths), aimed at generating collision-free paths for a set of agents from their initial position to their assigned goal on a given graph (often, a gridworld).
Most traditional centralized planning methods solve MAPF by following movement rules~\cite{sajid2012multi}, planing paths based on the priority order of agents~\cite{erdmann1987multiple}, expending the search space when a collision is found~\cite{wagner2015subdimensional}, or ignoring other agents first and resolving collisions afterwards~\cite{sharon2015conflict}.
These approaches are usually unable to scale to larger teams while maintaining high-quality solutions, and often struggle with the frequent online re-planning required in realistic deployments, due to their high computational cost.

Recently, the MAPF community has turned to Multi-Agent Reinforcement Learning (MARL) to produce fast and scalable, but suboptimal solutions to the MAPF problem by relying on decentralized planning under partial observability (to help scale to arbitrary world sizes).
However, this approach severely limit the amount of information available to the agents, making it difficult for them to detect and enact the type of joint maneuvers required to complete complex MAPF tasks.
Introducing (learned) communication to allow agents to share information within a team, thereby increasing each other's knowledge/observations, can help augment the amount of information available to single agent and promote team cooperation.
However, designing scalable communication learning mechanisms is highly nontrivial: as the team size increases, agents are more likely to be overwhelmed by (potentially contradictory) messages that essentially introduce noise into their observation (i.e., the \textit{chatter problem}).

This work extends our previous works, PRIMAL/PRIMAL2~\cite{sartoretti2019primal,damani2021primal}, which combine Reinforcement Learning (RL) and imitation learning to teach fully-decentralized policies for MAPF.
We propose SCRIMP, a new decentralized MAPF planner where agents make individual decisions based on their own, very small field of view (FOV) and on global highly-scalable communications\footnote{Note that an earlier version of this work was briefly described in an extended abstract (poster presentation) at AAMAS '23~\cite{aamas2023scrimp}.}.
The three main contributions of this work are:
First, we introduce a scalable and differentiable Transformer-based communication mechanism to effectively mitigate the hazards caused by partial observations and further promote team-level cooperation, while alleviating the chatter problem.
Second, we propose a learning-based stochastic tie-breaking method to handle scenarios in which agents' preferred actions would lead to \textit{vertex conflicts} (agents choose actions that would have them occupy the same cell at the next time step) or \textit{swap conflicts} (agents' actions would have them swap locations in a single time step), and efficiently break symmetries to maximize the agents' progression.
Third, we propose a simple intrinsic rewards structure, aimed at improving the agents' single-episode exploration, thus increasing their probability to reach their goal and mitigating the long-term credit assignment problem.
Throughout this work, to simplify implementation on robot that often rely on limited-range onboard sensing, we train agents with a small $3 \times 3$ FOV (the agent can only see itself and one grid cell around it).
We further show that the benefit from expanding this FOV is not significant, as the communication mechanism of SCRIMP already provides sufficient information (see Supplementary Material\footnote{Available at 
\url{https://drive.google.com/drive/folders/1wEm4p3jICdpoWf032egrcVSr9I3tMTIZ}}).
We test our trained models on an extensive set of simulation experiments with various team sizes, world sizes, and obstacle densities.
These evaluation results show that our models can scale nearly-losslessly to larger teams without retraining, and outperforms state-of-the-art learning-based MAPF planners (even with larger FOVs) in most relevant metrics.
In the majority of cases, our models actually perform similarly to the classical centralized planner ODrM*~\cite{wagner2015subdimensional}, and even achieve higher success rate in complex tasks that require joint maneuvers.
Our ablation studies further validate the effectiveness of the three proposed techniques.
We finally present simulation results of our trained policy implemented on robots in Gazebo.

\section{Related Work}

\subsection{Multi-Agent Reinforcement Learning}
\label{sec:MARL}

MARL approaches can be broadly classified into three big paradigms, namely Centralized Learning (CL), Independent Learning (IL) and Centralized Training
Decentralized Execution (CTDE)~\cite{wang2022distributed}.
CL methods model a multi-agent system as a single-agent and aim to learn a joint policy from the joint observations of all agents.
An obvious flaw is that the dimension of the state-action space increases exponentially with the number of agents, making learning hard and not scalable to larger teams.
In IL, each agent independently learns its own policy based on its own observation, while considering other agents as a part of the environment.
In partially observable environments, these individual policies usually tend to be strongly sub-optimal and unstable due to the lack of sufficient information, preventing agents from cooperating well.

By allowing agents to learn in a centralized setting while executing in a decentralized setting, the CTDE paradigm aims at getting the best of both CL and IL.
One effective approach in CTDE is to introduce explicit/implicit communication among agents (potentially learned as well), to share relevant information and/or intents, towards true joint cooperation at the team-level~\cite{shaw2022formic}\cite{zhang2019efficient}. 
In addition, based on shared additional information at training time, the learning speed and final performance of CTDE paradigm can be both improved through learning centralized critics~\cite{foerster2018counterfactual} or value factorization networks~\cite{rashid2018qmix}, etc.

\subsection{Reinforcement Learning-based MAPF}

Solving MAPF via MARL, often using demonstrations from a classical planner, has recently drawn great attention from the community.
The first work, named PRIMAL~\cite{sartoretti2019primal}, proposed to combine RL and imitation learning to plan paths using fully-decentralized policies in a partially observable environment.
The RL part is trained by the asynchronous advantage actor critic (A3C) algorithm, where all agents share the same parameters.
The imitation learning part relies on behavior cloning using training data generated online from a dynamically-coupled planner (ODrM*~\cite{wagner2015subdimensional}).
PRIMAL2~\cite{damani2021primal} then extended PRIMAL to lifelong MAPF, and relied on learned conventions to improve cooperation among agents in highly-structured environments.

More recent works have looked to communication learning as another promising direction to improve solution quality.
Similar to our work, MAGAT~\cite{li2021message} and DHC~\cite{ma2021distributed} both introduced Graph Neural Networks (GNN) based communication-learning by treating each agent as a node, and making decisions on aggregated information from its close-by neighbors.
DCC~\cite{ma2021learning} learns a request-reply selective communication by determining whether the central agent's decision is altered by the presence of a neighbor.
More recently, PICO~\cite{li2022multi} integrated planning priorities generated by a classical coupled planner into an ad-hoc routing communication topology to produce a collision-reducing policy.
However, all these communication learning based methods exhibit reduced scalability (usually cannot solve tasks with more than 64 agents) compared to earlier communication-free approaches, exhibiting a trade-off between solution quality and success rate as the team size increases.

\section{MAPF as a RL Problem}

\subsection{MAPF Problem Statement}

In this work, we consider MAPF problems composed of a set of $n$ agents $\left\{1, \ldots, n\right\}$, represented by a tuple $\langle G, S, T\rangle$, where $G=(V, E)$ is an undirected graph, $S=\left\{s_1, \ldots, s_n \right\} \subset V$ and $T=\left\{t_1, \ldots, t_n\right\} \subset V$ respectively contain $n$ unique start vertices and $n$ unique goal vertices, one for each agent~\cite{stern2019multi}.
Time is discretized into steps.
At each discrete time step, each agent (currently located at one of the graph's vertices) is allowed to execute a single action: either move to one of its adjacent vertices in the graph, or wait at its current vertex.
Movement actions are only feasible if they do not result in vertex/edge/swapping collisions.
The goal is to find the collision-free, shortest joint path that consists of $n$ single agent paths, such that all agents start from $S$, and finally end up at $T$ at some terminal time step, via a sequence of feasible actions.

\begin{figure*}[t]
  \centering
  \setlength{\belowcaptionskip}{-0.3cm}
  \includegraphics[width=0.8\linewidth]{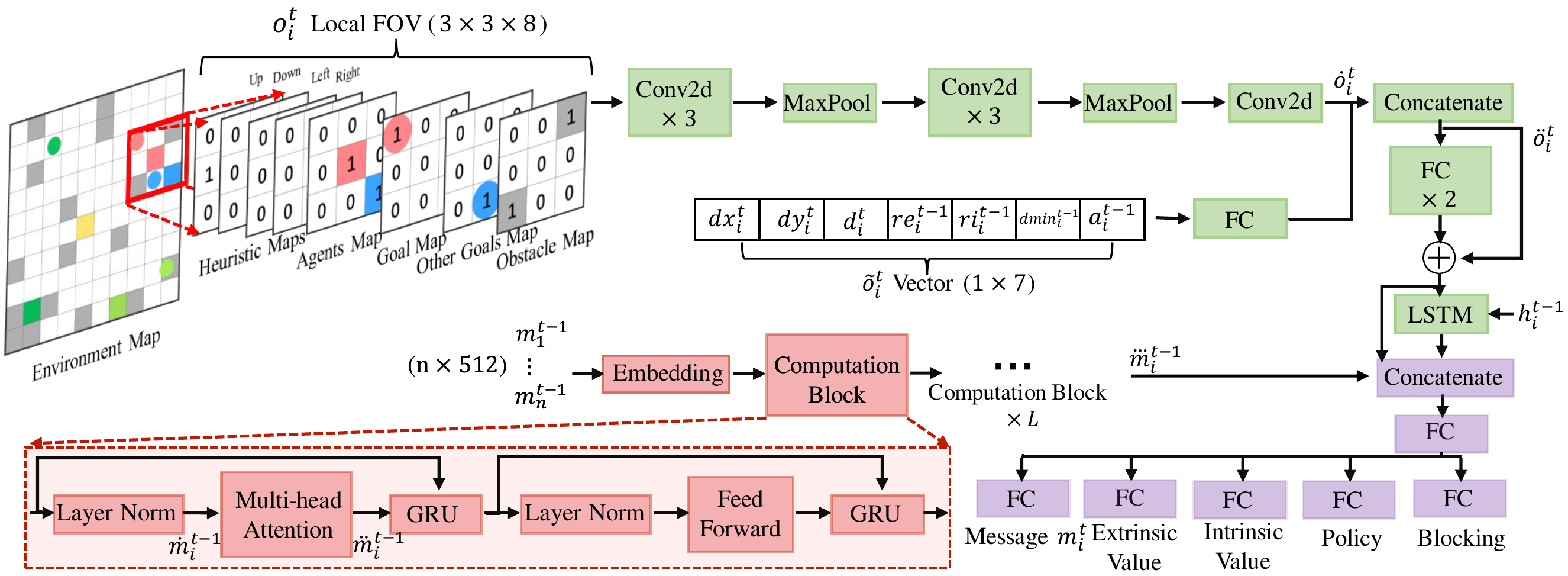}
  \vspace{-0.1cm}
  \captionsetup{font={small,bf,stretch=1}, justification=raggedright}
  \caption{Network structure of SCRIMP.
  Green modules belong to the observation encoder which is nearly-identical as PRIMAL, red modules to the transformer-based communication block, and purple modules to the output heads. 
  The transformer-based communication block consists of a stack of $L$ computation blocks. 
  The details of one computation block are shown in the red dashed box at the bottom of the figure.
  In the environment map, the gray squares represent static obstacles, and corresponding pairs of agents (squares) and goals (circles) are shown in the same color.}
  \label{fig:net}
  \vspace{-0.15cm}
\end{figure*}

\subsection{RL Environment Setup} 
\label{sec:env}

In line with the classical MAPF problem described above, we consider a discrete 2D gridworld environment where agents, goals, and obstacles each occupy one grid cell.
The world is described by a $m \times m$ matrix, where 0 represents an empty cell.
At the beginning of each episode, $n$ random, non-repeating start positions and $n$ corresponding goal positions are assigned to the $n$ agents.
In doing so, we also ensure that each agent's start and goal positions are randomly sampled from empty cells of the same connected region of the environment.
Different from previous work~\cite{sartoretti2019primal}, where agents are allowed to enact their action sequentially at each time step, our environment allows agents to move simultaneously and checks for collisions on this \textit{joint} action vector.
Such operation is closer to the standard MAPF setting used by centralized planners, and in particular allows agents to follow each other with no empty space between them.
As it is more realistic for real-world applications, we keep all of the agents in the environment after they have reached their goal.
The MAPF environment is set to be partially observable by only allowing each agent access to the information inside its FOV ($3 \times 3$ in our experiments).
This partial-observability assumption enables the trained policy to scale to arbitrary world sizes by fixing the dimensions of the neural network's inputs, and to remain closer to practical applications where robots will usually rely on onboard sensors whose view can be occluded by close-by obstacles/agents.

The observation of each agent is grouped into two parts.
The first part consists of eight binary matrices.
The first four heuristic maps correspond to each of the four movement actions, where a position is marked as 1 if and only if the agent would get closer to its goal by taking that action~\cite{ma2021distributed}.
The remaining four maps represent, respectively, nearby obstacles, the positions of other agents, the agent's own goal location (if within the FOV), and the projected position of other observable agents’ goals.
All locations outside the world's boundaries are regarded as obstacles.
The second part is a seven-length vector consisting of the current normalized distance between the agent and its own goal along the x-axis $dx$, y-axis $dy$, and total Euclidean distance $d$, extrinsic reward $re^{t-1}$ (see below and in Table~\ref{table:reward}), intrinsic reward $ri^{t-1}$ and minimum Euclidean distance between the agent's previous location and past locations stored in its episodic memory buffer $dmin^{t-1}$ (both described in Section~\ref{buffer}), as well as its most recent action $a^{t-1}$.

Our reward structure is shown in Table~\ref{table:reward}.
Aligned with previous works, agents currently off-goal are penalized at each step to encourage faster task completion.
An episode terminates if all agents are on goal at the end of a time step, or when the episode reaches the pre-defined time limit (in practice, 256 time steps).
Agents can communicate with each other, with a one time step delay from sending a message to receiving/using it and the communications are not hindered by obstacles.

\begin{table}[h]
\vspace{-0.5cm}
\centering
\captionsetup{font={small,bf,stretch=1}, justification=raggedright}
\caption{Reward structure}
\vspace{-0.1cm}
\label{table:reward}
\scalebox{1.}{
\begin{tabular}{|l|l|}
\hline
Action & Reward \\ \hline
Move(Up/Down/Left/Right) & -0.3 \\ \hline
Stay(On goal, Away goal) & 0.0,-0.3 \\ \hline
Collision & -2 \\ \hline
Block & -1 \\ \hline
\end{tabular}}
\vspace{-0.3cm}
\end{table}
\vspace{-0.2cm}

\section{Learning Techniques}

\subsection{Transformer-based Communication Learning}

To address the shortcomings of CL and IL described in Section~\ref{sec:MARL}, this work learns a common decentralized policy network while simultaneously introducing a scalable and differentiable communication mechanism that allows agents to gain knowledge about other agents' past observations, thereby effectively mitigating the risks posed by partial observations and further facilitating team-level cooperation that can not be learned under a communication-free decentralized framework.
Our communication mechanism is based on the transformer model~\cite{vaswani2017attention}, which has recently shown ground-breaking success in numerous fields of artificial intelligence.
The transformer's ability to efficiently integrate information over long sequences and scale to large amounts of data matches the primary requirement of our communications, which is to fuse messages sent by multiple agents into a single message in a scalable manner.
In addition, the transformer's ability to model dynamic pairwise relationships between messages without considering the order of inputs helps mitigate the \textit{chatter problem}.
A typical method in communication learning is to use the Long Short-Term Memory (LSTM) units as the communication channel.
Since transformers have demonstrated significant performance gains compared to LSTM in other domains, we hypothesize that transformer-based communication mechanisms should also outperform such LSTM-based communication mechanisms~\cite{wang2022fcmnet} in MARL.
However, in contrast to fields such as natural language processing, computer vision, etc., only a small amount of data is available in online RL for training models.
In addition, RL models must be able to rapidly adapt to changing tasks, which makes it challenging to optimize the classical transformer and often leads its performance to only remain comparable to stochastic strategies~\cite{mishra2017simple}.
To stabilize the training, we removed dropout layers, rearranged the location of layer normalizations in the submodules, and additionally added a Gate Recurrent Unit (GRU) in place of the residual connection of the vanilla transformer, as recommended in~\cite{parisotto2020stabilizing}.

Fig.~\ref{fig:net} shows the network structure of SCRIMP, composed of three modules: the observation encoder, the transformer-based communication block, and the output~heads.

\textit{a) Observation Encoder}: 
Seven convolutional layers, two max pooling layers, three Fully Connected (FC) layers, and one LSTM unit make up the observation encoder.
Specifically, for agent $i$ at time step $t$, the local observation $o_i^t$ consisting of a four-channel matrix is encoded into $\dot{o}_i^t$, via two stages of three convolutions and one maxpooling, followed by a last convolutional layer. 
In parallel, the vector of length seven $\tilde{o}_i^t$ is independently encoded by a FC layer and then concatenated with the matrix $\dot{o}_i^t$ to form $\ddot{o}_i^t$.
The encoded observation $\ddot{o}_i^t$ is finally passed through two FC layers and then fed into an LSTM unit together with the output hidden state $h_i^{t-1}$ of the unit at the previous time step, to endow each agent with the ability to integrate (its own) past information across time.

\textit{b) Transformer-Based Communication Block}:
Here, \textbf{only the encoder of the transformer is used}.
\textbf{It applies to different agents, not to time}, i.e., the transformer's encoder treats messages output by all agents at the previous time step $\left\{m_1^{t-1}, \ldots m_n^{t-1}\right\}$ as input.
These messages are first augmented with information about the unique ID of each agent using a sinusoidal positional embedding, and passed through multiple iterative computation blocks to produce the final output of the communication block $\dddot{m}_i^{t-1}$.
Each computation block includes a multi-head self-attention mechanism, a multi-layer perceptron, residual connections, layer normalizations, and GRUs.
The multi-head self-attention is the central block of our communication mechanism.
It encodes not only the high-level information of messages but also the dynamic interrelationships between messages.
For each agent $i$, at each of the $h$ independent attention head, its own message is linearly projected to a Query by learnable weights $w_q$\footnote{In practice, the attention function is calculated simultaneously over the set of all agents' queries, which are linearly projected by the weights $W_q$, packed into a matrix $Q$.}, while all messages are linearly projected to the Keys and Values by learnable weights $W_k$, $W_v$, and packed into matrix $K$ and $V$.
The relation between the self-message of agent $i$ and messages from all agents is then computed as:
\vspace{-0.35cm}
\begin{equation}
\mbox{\small\(
\alpha_{i}=\operatorname{softmax}\left(\frac{\boldsymbol{w}_q \dot{m}_i^{t-1} \cdot\left(\boldsymbol{W}_k\left[\dot{m}_1^{t-1}, \ldots \dot{m}_n^{t-1}\right]\right)^{\top}}{\sqrt{d_K}}\right),
\)}
\vspace{-0.2cm}
\end{equation}
\noindent where $\dot{m}_i^{t-1}$ is the representation of $m_i^{t-1}$ processed by the previous layers; $\sqrt{d_K}$ is the dimension of the Keys, used as a scaling factor to alleviate the gradient vanishing problem of the Softmax function.
The Values are weighted-summed over $\alpha_{i}$ to produce the final output of this attention head:

$\,$ \vspace{-0.4cm}
\begin{equation}
\text {head}_{i}=\alpha_{i} \boldsymbol{W}_v\left[\dot{m}_1^{t-1}, \ldots \dot{m}_n^{t-1}\right].
\vspace{-0.16cm}
\end{equation}
The output of all heads are then concatenated and projected by the learnable weight $W_o$ to obtain the final joint output from different representation subspaces, that is
\vspace{-0.18cm}
\begin{equation}
\ddot{m}_i^{t-1}=\text { Concat }\left(\text { head }_{i_1}, \ldots, \text { head }_{i_h}\right) W_o.
\vspace{-0.12cm}
\end{equation}
When global communication is not available, SCRIMP can be extended to a local communication version that limits the communication range based on Euclidean distance.
The local communication is achieved by masking (setting to $-\infty$) inputs of the Softmax function from illegal communication in the dot product attention.
Since the encoder is placed before the output heads, agents' decisions are conditioned on their own observation as well as the messages exchanged within the team.
Because the encoder is differentiable, it is trained using the gradient of all agents' losses, and as training proceeds, it learns to capture more useful information and to ignore noise/mitigate chatter to reduce the losses of all agents, leading to a policy that exhibits enhanced and larger-scale cooperation.

\textit{c) Output Heads}:
Finally, the output of the transformer $\dddot{m}_i^{t-1}$ is concatenated with the output and the input of the LSTM cell, which is then used by the output heads to generate 1) a message used at the next time step, 2) predicted state values for extrinsic and intrinsic reward (detailed in Section~\ref{buffer}), 3) the agent's policy for the current time step, and 4) the learned ``blocking'' output used to indicate whether the agent is blocking other agents from reaching their goals~\cite{sartoretti2019primal}.
Our model is trained by proximal policy optimization (PPO)~\cite{schulman2017proximal}.
During training, the predictions of state values for extrinsic and intrinsic reward are both updated by minimizing the temporal difference (TD) error between the predicted value and the discounted return.
The output policy is optimized using the loss:
\vspace{-0.2cm}
\begin{equation}
\mbox{\small\(
\begin{split}
L_\pi(\theta)&=\frac{1}{T} \frac{1}{n} \sum_{t=1}^T \sum_{i=1}^n \min \left(r_i^t(\theta) \widehat{A_i^t}\right., \\ &\quad \left.\operatorname{clip}\left(r_i^t(\theta), 1-\epsilon, 1+\epsilon\right) \widehat{A_i^t}\right),
\end{split}\)}
\vspace{-0.15cm}
\end{equation}
where $\theta$ denotes the parameters of the neural network, $r_i^t(\theta)$ the clipped probability ratio $r_i^t(\theta)=\frac{\pi_i^\theta\left(a_i^t \mid o_i^t, \tilde{o}_i^t, h_i^{t-1}, m_1^{t-1} \cdots, m_n^{t-1}\right)}{\pi_i^{\theta_{\text {old }}}\left(a_i^t \mid o_i^t, \tilde{o}_i^t, h_i^{t-1}, m_1^{t-1} \ldots, m_n^{t-1}\right)}$ with $\pi_i^\theta$ and $\pi_i^{\theta_{\text {old }}}$ the new and old policies of the agent $i$, respectively.
$\epsilon$ is a clipping hyperparameter, and $\widehat{A}_i^t$ is the truncated version of the generalized advantage function.
The "blocking'' prediction is trained using a binary cross-entropy loss, and implicitly helps agents understand the additional blocking penalty they might receive (see Section~\ref{others} for details).
The total loss also includes an entropy term, which has been shown to encourage exploration and inhibit premature convergence.

\begin{figure}[t]
  \vspace{0.15cm}
  \centering
  \setlength{\belowcaptionskip}{-0.3cm}
  \includegraphics[width=0.25\linewidth]{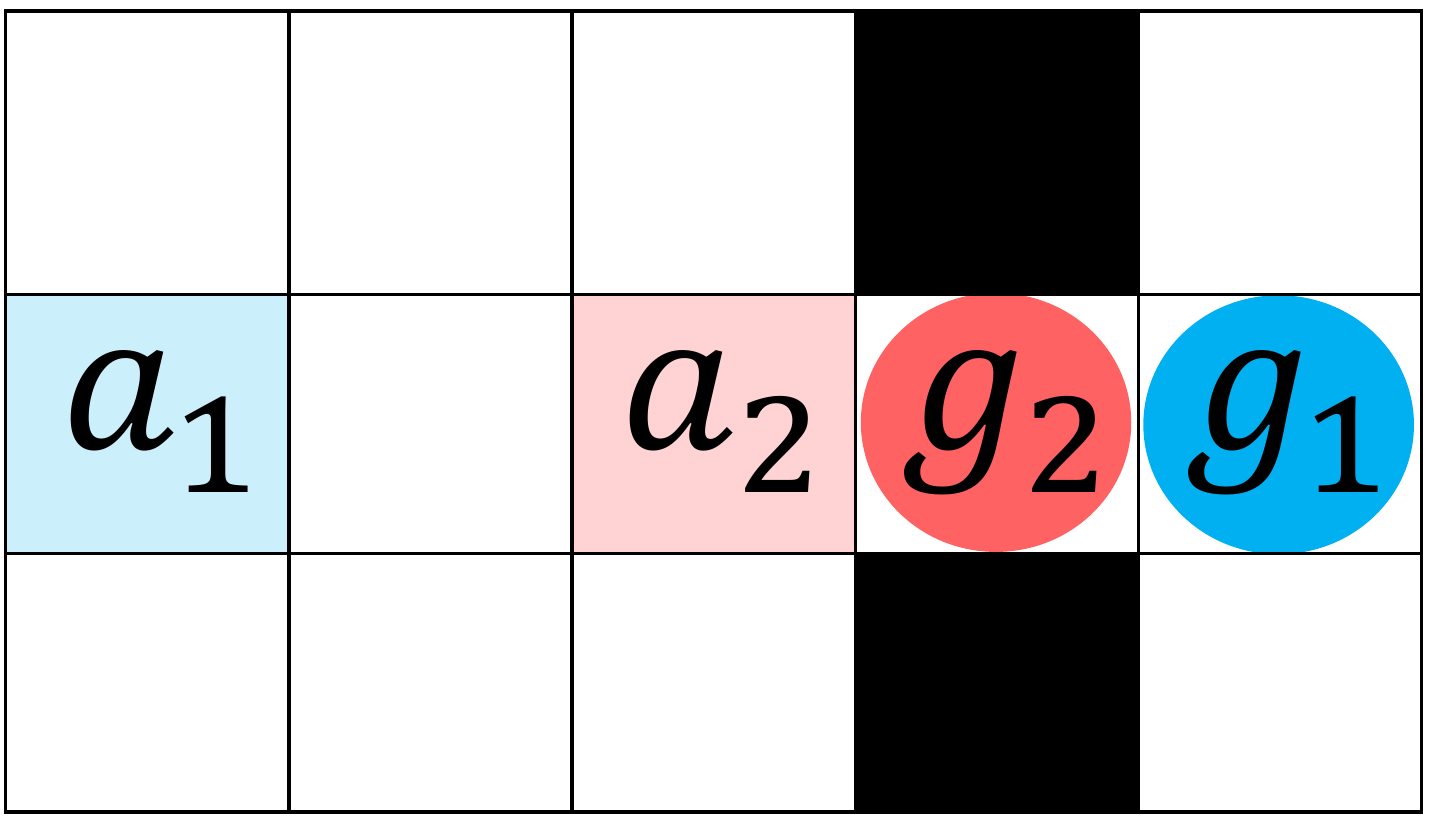}
  \vspace{-0.1cm}
  \captionsetup{font={small,bf,stretch=1}, justification=raggedright}
  \caption{Example of target symmetry. Agent $1$ ($a_1$) needs to traverse the goal ($g_2$) of $a_2$ to reach its own goal ($g_1$).}
  \label{fig:conflict}
  \vspace{-0.3cm}
\end{figure}

\subsection{Value-Based Tie Breaking}

Inter-agent collisions/deadlocks is one of the most important causes of low-quality paths, and is the main challenge in complex MAPF tasks.
We expect dramatic gains in performance if agents can reach a meaningful consensus on the priority of competing moves before a conflict occurs, thereby effectively breaking such symmetries.
Previous works often proposed to handle vertex or swap conflicts, by discarding these actions and instead letting agents remain at their current position and receive a large penalty.
In this work, we propose to let agents check if they will collide at the next time step before the action is actually executed, and then (weighted-)randomly sample which agent is finally allowed to perform its preferred action, based on a learning-based \textit{priority probability}.
Other agents then re-sample a new action according to their policy, where their former, conflicting choice is masked.
We propose to use the predicted difference in team state-value as a component in the calculation of the priority probability, since it represents the long-term collective benefit of that agent being allowed to choose this specific action, thus preventing selfish, short-sighted behaviors.
Specifically, if agent $i$ would collide with agent $j$ at the next step, we compute the difference between the predicted current team state value and the next team state value where agent $i$'s action is executed and agent $j$ re-samples a new action.
We also calculate the same difference for agent $j$ using the same process.
This difference in team state values for agent $i$ is defined as:
\vspace{-0.26cm}
\begin{equation}
\label{move diff}
\operatorname{diff}_i^t=\sum_{k=1}^n\left(\bar{v}_k^t-\bar{v}_{k_i}^{t+1}\right),
\vspace{-0.17cm}
\end{equation}
where $\bar{v}_k^t$ is the predicted state value of agent $k \in\{1, \ldots, n\}$ at step $t$, and $\bar{v}_{k_i}^{t+1}$ is that of agent $k \in\{1, \ldots, n\}$ at step $t+1$ obtained by agent $i$ performing the original action and agent $j$ re-sampling a new action.
Here, we assume that the team state value is the sum of the individual state value functions, because it is easy to implement and guarantees a global argmax operation on the team state value produces the same result as a collection of individual argmax operations on each individual state value~\cite{sunehag2017value}.

It is worth noting that an agent closer to its goal has a greater self-benefit from performing its preferred action, since fewer time steps (i.e., exponents of the discount factor) are required to obtain the higher rewards associated with reaching and staying on goal.
However, in the ``target symmetry'' scenario~\cite{li2020new} depicted in Figure~\ref{fig:conflict}, where one agent needs to move past another agent's goal to reach its own, the agent closer to its goal should be assigned a lower priority probability to prevent deadlocks, which may contradict the result calculated in Eq.~\eqref{move diff}.
Therefore, we introduce a second component in the calculation of the priority probability, which relies on the normalized Euclidean distance between the current position of each agent and its goal.
The weighted sums of the state-value differences and the normalized distances are fed into a softmax layer to generate the final priority probability:
\vspace{-0.18cm}
\begin{equation}
\mbox{\small\(
P=\operatorname{softmax}\left(\operatorname{diff}_i^t+\mu \frac{d_i^t}{\sum_k d_k^t}, \operatorname{diff}_j^t+\mu \frac{d_j^t}{\sum_k d_k^t}, \cdots\right), k \in\{i, j \ldots\},
\vspace{-0.05cm}
\)}
\end{equation}
where $\mu$ is a pre-defined hyperparameter to control the importance of distance, $d_i^t$ is the distance between agent $i$'s current position and its own goal, $\{i, j \ldots\}$ is the set of collided agents.
Finally, the agent allowed to move is sampled based on this probability.
We experimentally observed that always greedily selecting the highest priority probability agent as the movable agent gives slightly worse performance compared to this stochastic method.

\begin{table*}[t]
\centering
\captionsetup{font={small,bf,stretch=1}, justification=raggedright}
\caption{Average performance and standard deviation along standard MAPF metrics of different algorithms for instances with different team sizes, environment sizes, and obstacle densities.
All results are averaged over 100 episodes with randomly generated maps.
The episode length (EL) is the number of time steps required to complete the task.
Episode lengths for failed tasks (no solution after 5 minutes of planning for ODrM*, more than 256 time steps for learning-based methods) are excluded.
The maximum number of goals reached (MR) indicates the maximum number of on-goal agents at any point during an episode.
The collision ratio with static obstacles including boundaries (CO) is calculated as $\frac{\textbf{Number of collisions}}{\textbf{Episode length} \times \textbf{Number of agents}} \times 100 \%$.
The success rate (SR) is the percentage of episodes fully solved (i.e., where all agents reached their goal within the planning time/episode length constraints).
↑ indicates that ``higher is better'', and ↓ ``lower is better''.
"-" represents unavailable data.
The results of the best-performing algorithms in each task, except for ODrM*, are highlighted in bold.
}
\vspace{-0.2cm}
\label{table:performance}
\scalebox{0.65}{
\begin{tabular}{|l|lll|lll|lll|lll|}
\hline
\multicolumn{1}{|c|}{\multirow{2}{*}{Methods}} & \multicolumn{12}{c|}{8 agents in $10\times10$ world with 0\%, 15\%, 30\% static obstacle rate} \\ \cline{2-13} 
\multicolumn{1}{|c|}{} & \multicolumn{3}{c|}{EL ↓} & \multicolumn{3}{c|}{MR ↑} & \multicolumn{3}{c|}{CO ↓} & \multicolumn{3}{c|}{SR ↑} \\ \hline
ODrM* &12.64(2.13)  &  13.72(2.46)& 16.16(4.03) & - & - & - & 0.00\%(0.00\%) & 0.00\%(0.00\%) & 0.00\%(0.00\%) &100\%  & 100\% & 100\% \\
DHC &14.34(5.08) & 17.23(5.96) &29.82(20.93) & \textbf{8.00(0.00)} &7.98(0.14)  &7.88(0.50)  & \textbf{0.00\%(0.00\%)} &\textbf{0.00\%(0.00\%)} &\textbf{0.00\%(0.00\%)} &\textbf{100\%} & 98\% & 92\% \\
PICO &17.35(12.39)  & 29.23(28.71) & 35.04(26.93) &\textbf{8.00(0.00)}  &7.51(0.72) &6.68(1.76) &\textbf{0.00\%(0.00\%)}  &\textbf{0.00\%(0.00\%)} & \textbf{0.00\%(0.00\%)} &\textbf{100\%}  &63\% & 29\% \\
SCRIMP &\textbf{12.05(2.53)}&\textbf{14.17(3.52)}& \textbf{18.56(7.80)}  & \textbf{8.00(0.00)} & \textbf{8.00(0.00)} & \textbf{7.93(0.45)} & \textbf{0.00\%(0.00\%)} &0.04\%(0.18\%)  & 0.07\%(0.26\%)&\textbf{100\%}  & \textbf{100\%} & \textbf{97\%}\\\hline
 
 & \multicolumn{12}{c|}{32 agents in $30\times30$ world with 0\%, 15\%, 30\% static obstacle rate} \\ \hline
ODrM* & 42.97(4.62) & 43.39(4.97) & 53.95(10.43) & - & - & - & 0.00\%(0.00\%) & 0.00\%(0.00\%) & 0.00\%(0.00\%) &100\%  &100\%  & 97\% \\
DHC &48.64(17.54)&52.48(7.10) &91.29(29.25) &\textbf{32.00(0.00)}  & 31.97(0.22) & 31.61(1.10) & \textbf{0.00\%(0.00\%)} & \textbf{0.00\%(0.00\%)}&\textbf{0.00\%(0.00\%)} &\textbf{100\%}&98\%  & 78\% \\
PICO &65.14(10.30)&- &- &30.26(1.18)  &22.70(2.40) &12.20(2.53)  &\textbf{0.00\%(0.00\%) } &\textbf{0.00\%(0.00\%)} &\textbf{0.00\%(0.00\%)} &14\%&0\%  &0\%  \\
SCRIMP &\textbf{42.59(4.87)} &\textbf{43.90(5.11)} &\textbf{62.31(17.86)} &\textbf{32.00(0.00)} & \textbf{32.00(0.00)} & \textbf{31.94(0.34)} &\textbf{0.00\%(0.00\%)}  & 0.00\%(0.02\%)& 0.32\%(0.44\%)&\textbf{100\%} &\textbf{100\%} & \textbf{96\%}\\ \hline

 & \multicolumn{12}{c|}{128 agents in $40\times40$ world with 0\%, 15\%, 30\% static obstacle rate} \\ \hline
ODrM* & 64.55(4.97) & 67.77(6.64) & 76.90(7.12) &-  & - & - & 0.00\%(0.00\%) & 0.00\%(0.00\%) & 0.00\%(0.00\%) & 100\% & 100\% & 20\%  \\
DHC &93.00(39.24)&137.76(47.69) &- &127.92(0.31)  &127.45(1.06)  &97.83(16.08) & \textbf{0.00\%(0.00\%)} &\textbf{0.00\%(0.00\%)} & \textbf{0.00\%(0.00\%)}&93\%&70\%  & 0\% \\
PICO &135.00(84.85)&- &- & 126.59(3.88) & 95.01(6.97) &58.11(8.79)  & \textbf{0.00\%(0.00\%)} & 0.024\%(0.022\%)& 7.80\%(8.52\%)&2\%& 0\% & 0\% \\
SCRIMP& \textbf{66.86(5.83)}&\textbf{73.15(8.53) }&\textbf{135.43(42.32)} & \textbf{128.00(0.00)}&\textbf{128.00(0.00)}&\textbf{121.95(13.16)} & 0.01\%(0.01\%) & 0.14\%(0.13\%) & 3.72\%(4.40\%) & \textbf{100\%}& \textbf{100\%} & \textbf{58\%} \\\hline
\end{tabular}}
\vspace{-0.4cm}
\end{table*}

\subsection{Episodic Buffer for Improved Exploration}
\label{buffer}

Efficient exploration of the state-action space remains one of the main challenges in (MA)RL.
In general, strategies that become greedy too early during training often cannot learn to act optimally; they may converge prematurely to worse suboptimal policies and then never collect new data that would allow them to further improve~\cite{yang2021exploration}.
This is even more challenging in MAPF tasks, where agents may need hundreds of (lowly-rewarded) actions before they get any more significant signal for reaching and staying on their goal.
In practice, we often observe that agents whose goal is far from their current location may choose to safely wander around the same area in a safe, collision-free manner, rather than try to reach their goal by exploring new paths/strategies.
Recent years have seen considerable efforts to encourage exploration from different perspectives~\cite{amin2021survey}.
A common solution is to let agents create bonus rewards for themselves based on their familiarity with visited states.
These bonuses (i.e., \textit{intrinsic rewards}) are summed with their true task rewards (i.e., \textit{extrinsic} rewards), and let the RL algorithm learn from these combined rewards.
Most existing approaches measure state familiarity by building a model that predicts dynamic environmental properties, and devise intrinsic rewards proportional to the error or uncertainty in the model prediction~\cite{badia2020agent57}.
However, since our MAPF environment is a gridworld, we propose a simple and straightforward method that shares the same intuition as these approaches, but avoids the expensive resources required to build such a model, by generating intrinsic rewards based on the $(x,y)$ coordinates of the visited grid cells.

Rather than encouraging agents to simply visit more state-action pairs during training, our proposed approach focuses on encouraging agents to explore more new area within each episode, to increase their probability of reaching their goal.
We let each agent keep an independent buffer with a limited capacity $M$, which starts each episode empty.
At each step, each off-goal agent first separately calculate the Euclidean distance between its current $(x,y)$ location and all stored locations in the buffer.
If the maximum distance is greater than a threshold $\tau$, the agent generates an intrinsic reward:
\vspace{-0.16cm}
\begin{equation}
r i_i^t=\varphi(\beta-\delta), \quad \delta=\left\{\begin{array}{l}
1 \text {, if max distance }<\tau \\
0 \text {, if max distance } \geq \tau,
\end{array}\right.
\vspace{-0.1cm}
\end{equation}
where $\tau$, $\varphi \in \mathbb{R}^{+}$ and $\beta \in \{0,1\}$ are pre-defined hyperparameters.
Essentially, $\tau$ checks whether the agent has stepped out of the qualitative region it was familiar with so far during the current episode.
The value of $\varphi$ depends on the scale of the task reward, while the value of $\beta$ determines the sign of the intrinsic reward.
In our tasks, we set $\varphi=0.2$ and $\beta=1$, to ensure that the best total reward can still only be obtained when agents reach their goal as fast as possible and then stay on it.
The final reward is calculated as $r_i^t=ri_i^t+re_i^t$.
After calculating any intrinsic rewards, each agent determines whether its current maximum distance is greater than or equal to another threshold $\rho$; if so, the current $(x,y)$ location is added to the episodic buffer~\cite{savinov2018episodic}.
There, if the buffer is at full capacity $M$, a random element in the buffer is instead replaced with the current element.
The above process continues until the end of the episode, where the buffer is emptied and a new episode begins.

Aligned with famous single-agent exploration works such as Agent57~\cite{badia2020agent57}, we let the policy network use $ri_i^{t-1}$,$re_i^{t-1}$,$dmin_i^{t-1}$ as additional inputs, both during training and execution, and further train two separate value outputs to predict $vi_i^t$ and $ve_i^t$, which are summed into the final critic value.
In practice, we start giving intrinsic rewards only after agents have learned the basics of the task (empirically set to be after $10^6$ time steps).
Our experimental results show that these additional techniques bring improvements in training stability and final performance.

\subsection{Other Training Details}
\label{others}

In addition to the three learning techniques proposed above, we also keep the following techniques, which were introduced and shown to be effective in our previous works~\cite{damani2021primal}~\cite{sartoretti2019primal}.
At the beginning of each episode, we randomly choose whether the network will be trained by imitation learning (behavior cloning based on the centralized planner ODrM*) or RL, depending on a predetermined ratio.
We also rely on a (supervised) binary cross-entropy loss to teach agents to avoid undesirable actions (e.g., actions that cause an agent to collide with a static obstacle, including world boundaries, or to directly return to their previous location), since it helps obtain better solutions compared training this from negative rewards.
We further included a blocking penalty in our general reward structure to encourage agents to temporarily move away from their goal in specific selfish ``blocking'' situations, defined as the presence of an agent extending the A*-path length of another agent by more than 10 steps.
However, unlike our prior works~\cite{damani2021primal}\cite{sartoretti2019primal}, where agents only draw actions from the set of valid actions during training, in all experiments in this paper, actions are sampled from the full action space, as we find no significant performance reduction when doing so, and the training curves are thus easier to compare.

\section{Experiments}

\begin{figure*}[t] 
\centering
\subfigure[Communication Ablation] { \label{fig:comm_ablation}
\includegraphics[width=0.4\columnwidth]{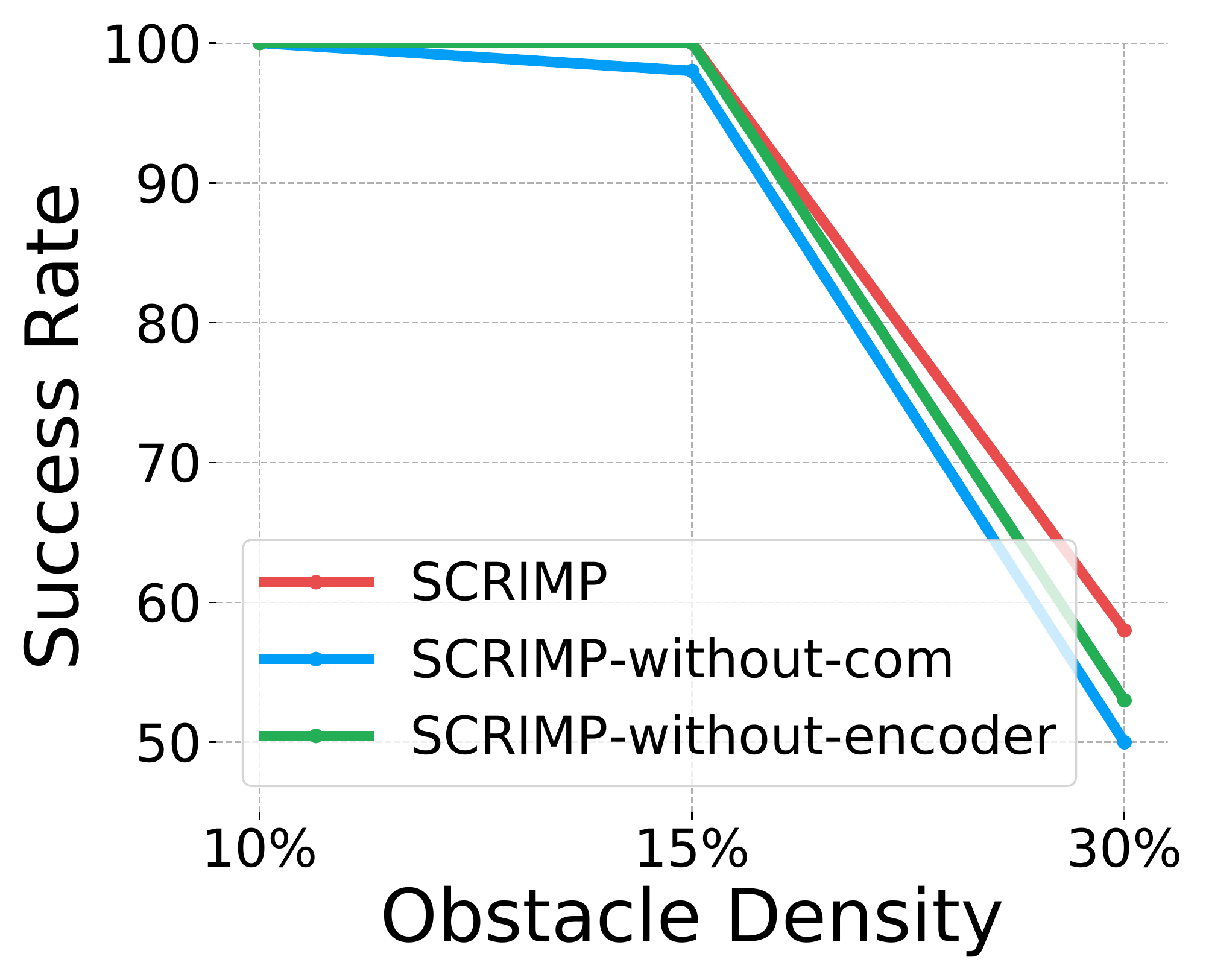}
}
\subfigure[Tie Breaking Ablation] { \label{fig:breaking_ablation}
\includegraphics[width=0.4\columnwidth]{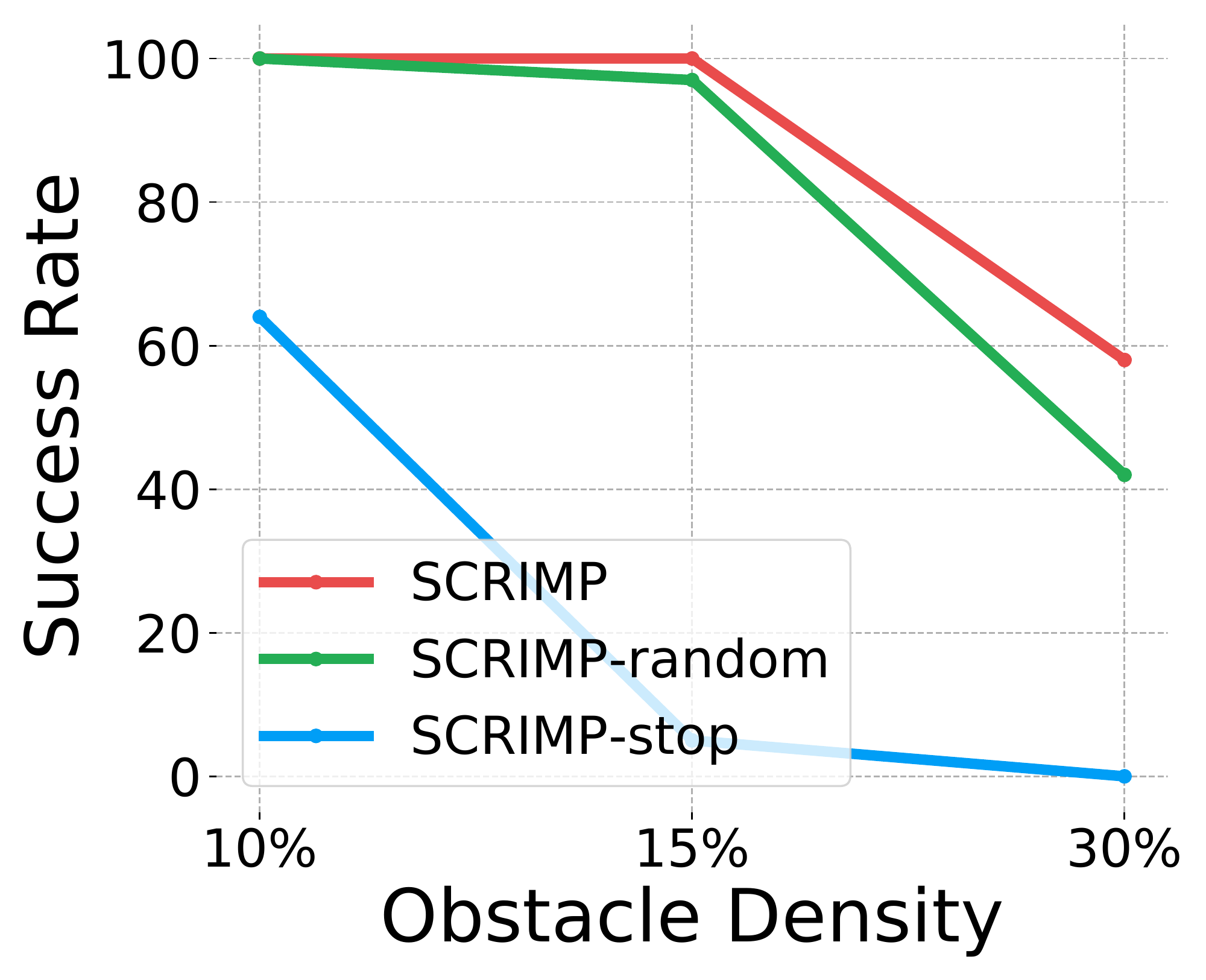}
}
\subfigure[Intrinsic Reward Ablation] { \label{fig:reward_ablation}
\includegraphics[width=0.4\columnwidth]{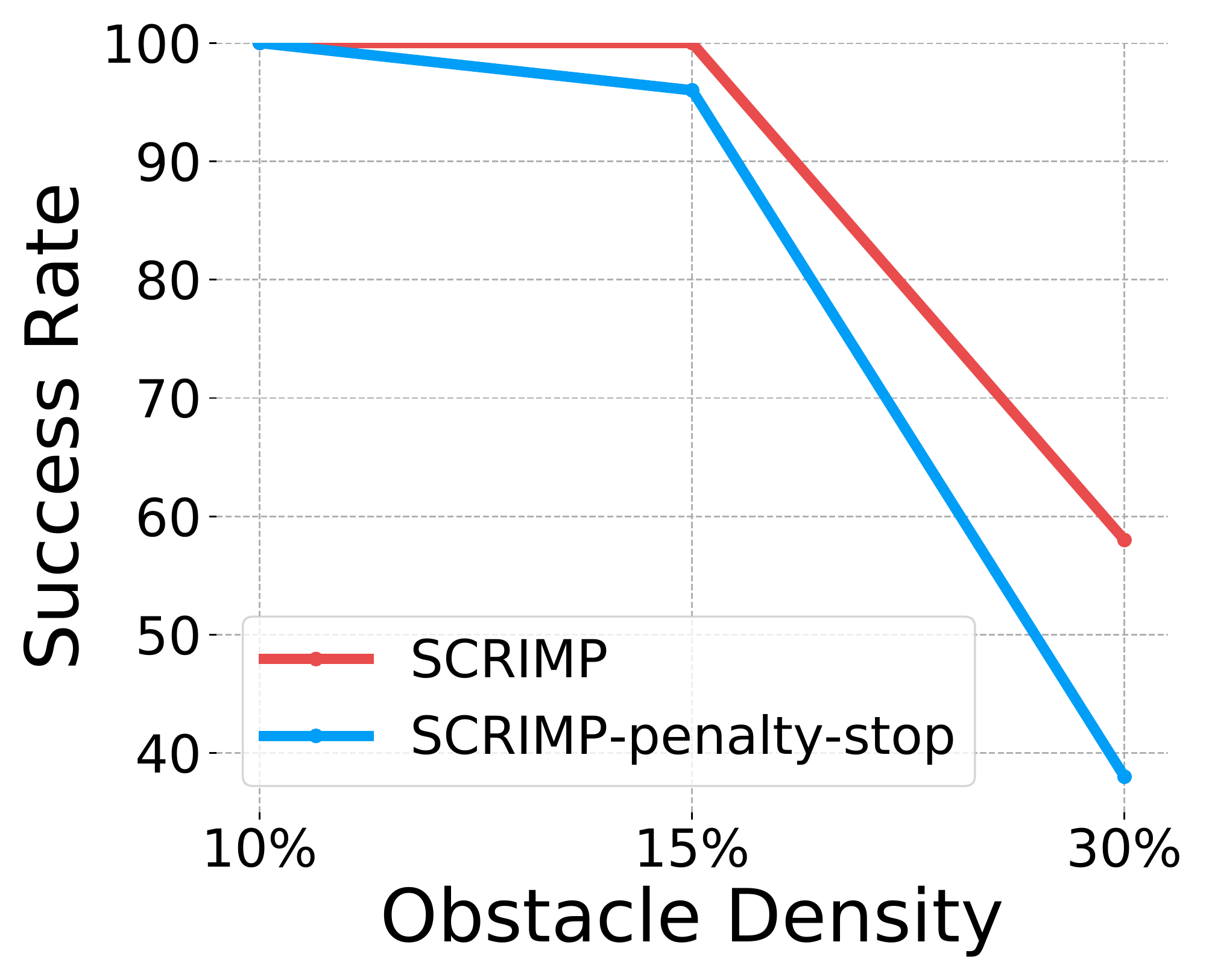}
}
\vspace{-0.3cm}
\captionsetup{font={small,bf,stretch=1}, justification=raggedright}
\caption{
Success rate in our ablation study.
The number of agents is equals to 128, $40\times40$ world size, and increasing static obstacle rates.}
\label{fig}
\vspace{-0.4cm}
\end{figure*}

\subsection{Comparison with Other MAPF Planners}
\label{performance experiment}
Our models are trained and tested in the MAPF environments described in Section~\ref{sec:env}.
During training, at the beginning of each episode, the size of the square gridworld is randomly selected as either $10$, $25$, or $40$.
The obstacle density is randomly selected from a triangular distribution between $0\%$ and $50\%$, with the peak located at $33\%$ (same as the compared baseline algorithms).
The number of agents is fixed at 8.
During testing, the number of agents varies among $8$, $32$, and $128$ and the gridworld size is varied from $10$ to $40$, the obstacle density varies among $0\%$, $15\%$, and $30\%$.
The experimental setup, additional evaluation results and the hyperparameters of our models are detailed in the Supplementary Material.
Our full training and testing code is available at \url{https://github.com/marmotlab/SCRIMP}.

We compare our method with two other state-of-the-art MARL methods, namely the GNN-based communication method DHC, and the ad-hoc routing communication method PICO, in terms of episode length to completed a task (\textbf{EL}), maximum number of goals reached (\textbf{MR}), collision rate with static obstacles including boundaries (\textbf{CO}), and success rate (\textbf{SR}).
All training hyperparameters of these two methods are the same as their official codes/paper, where the FOV of DHC is $9 \times 9$ and PICO is $11 \times 11$.
We also show the results of the bounded-optimal centralized planner ODrM* with inflation factor $\epsilon= 2.0$ and a timeout of $5$ minutes as a reference.

Testing results are presented in Table~\ref{table:performance}.
Overall, SCRIMP outperforms the other two learning-based methods, DHC and PICO, in almost all cases in terms of EL, MR and SR.
Furthermore, SCRIMP is even able to yield similar performance as the classical centralized planner ODrM*, even though ODrM* has access to the entire state of the system and relies on (at worst, exhaustive) search-based planning, and can guarantee bounded-optimal solutions given sufficiently large planning times.
In particular, SCRIMP still achieves 58\% SR in the most difficult task requiring a high degree of agent cooperation (128 agents, 30\% obstacle density), while the SR of ODrM* is only 20\% within 5min of planning, and DHC and PICO are completely unable to solve the task.
Although the CO of our models is already very low, in most cases it is higher than the CO of DHC and PICO.
A possible reason for this is that our intrinsic rewards encourage SCRIMP to learn less conservative policies to approach goals, where the probability of encountering obstacles in agents' long-term movement path inevitably rises.
In contrast, the policies of the other two approaches tend to keep agents within their starting region more often, and thus their EL, MR and SR are significantly worse than our models despite having a lower CO.

\subsection{Ablation Results}

\subsubsection{Communication Mechanism Ablation}

To fairly compare the importance of our Transformer-based communications, we develop two ablation baselines: SCRIMP-without-com and SCRIMP-without-encoder.
The structures of these SCRIMP variants are identical to the original SCRIMP, except that for SCRIMP-without-com, the entire communication mechanism is completely removed, i.e., agents never receive/use messages.
For SCRIMP-without-encoder, we only remove the transformer encoder and replace the fused message in the original SCRIMP with a self-message, i.e., agents cannot communicate with each other but are able to receive a message from themselves from the previous step.
Our evaluation results indicate that in simple tasks with fewer than 128 agents, SCRIMP performs similarly to these two variants, with success rates close to 100\% in most tasks.
However, in challenging tasks, the advantage of SCRIMP start to emerge as shown in Figure~\ref{fig:comm_ablation}.
SCRIMP-without-com exhibits the lowest success rate, which indicates that the content being encoded is essential for good final performance.
SCRIMP-without-encoder also performs worse than SCRIMP.
We believe that these results validate the effectiveness of obtaining information about other agents through dynamic fusion in harder tasks that are highly dependent on agent cooperation.

\subsubsection{Tie Breaking Ablation}

In this ablation experiment, when a inter-agent collision is about to occur, SCRIMP-stop uses the same processing as PRIMAL, i.e., all agents that cause the collision stop at their previous position and receive a $-2$ penalty.
SCRIMP-random follows a similar strategy to the original SCRIMP, but where the agent allowed to move at a conflict is randomly determined.
Figure~\ref{fig:breaking_ablation} demonstrates that allowing agents to re-sample actions instead of directly stopping yields great benefits, as expected, but if the movement priority is completely randomly determined, unknown instabilities will deteriorate the performance of the final learned policies.

\subsubsection{Intrinsic Reward Ablation}

Finally, we compare SCRIMP with SCRIMP-penalty-stop.
This variant removes the intrinsic reward completely (including the additional network inputs and output).
Same as in PRIMAL, we now give a slightly higher negative reward to an agent stopping off-goal ($-0.5$) than for moving ($-0.3$) to implicitly encourage exploration.
The SR of SCRIMP-penalty-stop and SCRIMP are also similar on simple tasks, but the gap between the baseline and SCRIMP on complex tasks, as shown in Figure~\ref{fig:reward_ablation}, suggests that our proposed approach effectively encourages exploration while balancing exploitation.

\subsection{Experimental Validation}

\begin{figure}[t]
  \vspace{-0.3cm}
  \centering
  \setlength{\belowcaptionskip}{-0.3cm}
  \includegraphics[width=0.6\linewidth]{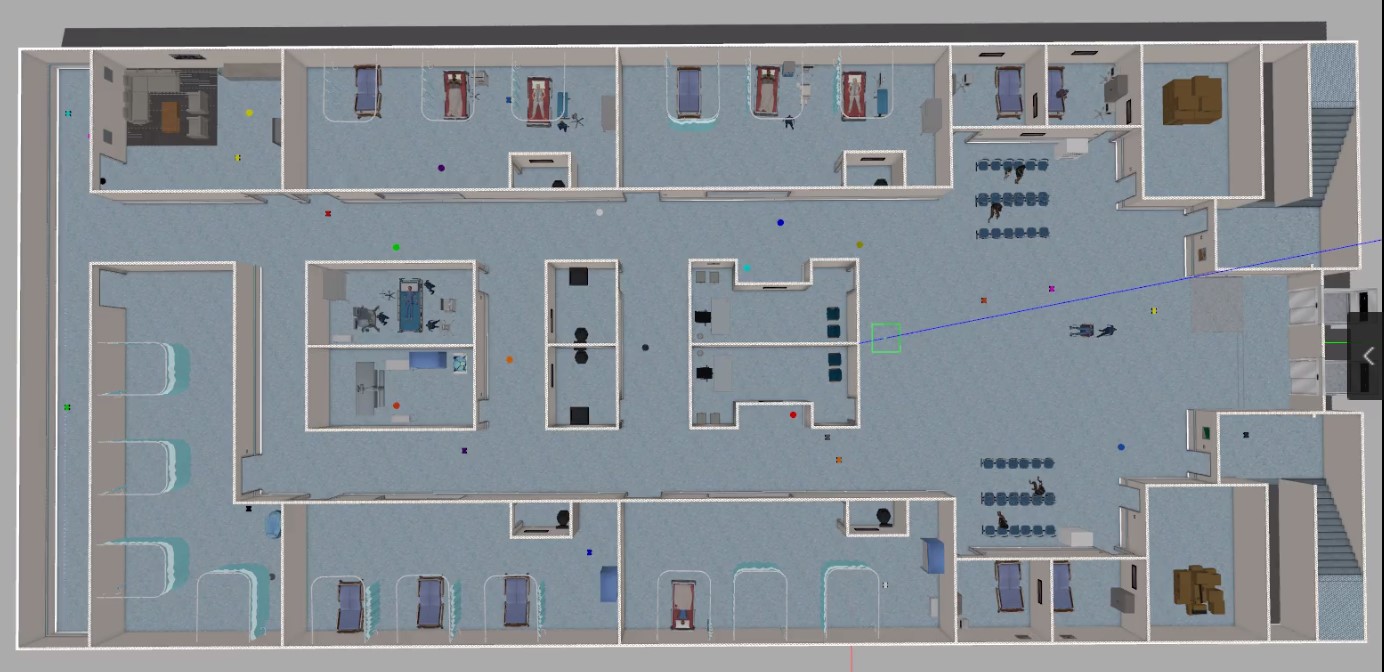}
  \vspace{-0.1cm}
  \captionsetup{font={small,bf,stretch=1}, justification=raggedright}
  \caption{Gazebo Environment setup. Sixteen autonomous ground vehicles and their corresponding goals in a high structured simulation environment.}
  \label{fig:simulation}
  \vspace{-0.3cm}
\end{figure}

We implemented our trained SCRIMP model on a team of sixteen autonomous ground vehicles in Gazebo, without any additional training/modification.
Figure \ref{fig:simulation} shows our highly structured simulation environment. 
Although the online planner is trained in a randomised grid world with no clear spatial structure, all vehicles are able to reach their goals quickly without collision.
The full video is available in the Supplementary Material.

\section{Conclusion}

This work introduces SCRIMP, a new RL approach to MAPF that relies on a highly scalable communication mechanism.
Our proposed communication learning method utilizes a modified transformer encoder to selectively and dynamically fuse information from other agents, enabling effective information sharing while avoiding overwhelming communications, thus significantly reducing the risk caused by partial observations and enabling cooperative operations that cannot be learned by a fully decentralized RL approach.
We further propose to break ties between competing agents online, by using a stochastic priority-based system based on learned, long-term team values.
Additionally, SCRIMP employs intrinsic rewards based on an episodic buffer, which encourages agents to explore more areas within an episode, improving the probability of reaching the goal and alleviating the long-term credit assignment issue.
Our experimental results across a range of tasks show that SCRIMP significantly outperforms other learning-based state-of-the-art planners, while maintaining scalability to larger teams.
In most cases, our models even perform similarly to a conventional search-based centralized planner, and even achieves higher success rate in our largest, most complex task.
Through a set of ablation studies, we demonstrate the effectiveness of the three proposed methods.
Finally, we implement our trained model on robot in a high-fidelity simulator, showing that robots can rely on our learned policy to plan online collision-free paths in real-time.
Future work will extend our approach to more practical MAPF tasks where each agent's vision can be obstructed by obstacles, and focus on environments that may contain dynamic obstacles such as humans.

\section*{ACKNOWLEDGMENT}

This work was supported by the Singapore Ministry of Education Academic Research Fund Tier 1.

\bibliographystyle{IEEEtran} 
\bibliography{root}

\begin{thebibliography}{10}
\providecommand{\url}[1]{#1}
\csname url@rmstyle\endcsname
\providecommand{\newblock}{\relax}
\providecommand{\bibinfo}[2]{#2}
\providecommand\BIBentrySTDinterwordspacing{\spaceskip=0pt\relax}
\providecommand\BIBentryALTinterwordstretchfactor{4}
\providecommand\BIBentryALTinterwordspacing{\spaceskip=\fontdimen2\font plus
\BIBentryALTinterwordstretchfactor\fontdimen3\font minus
  \fontdimen4\font\relax}
\providecommand\BIBforeignlanguage[2]{{%
\expandafter\ifx\csname l@#1\endcsname\relax
\typeout{** WARNING: IEEEtran.bst: No hyphenation pattern has been}%
\typeout{** loaded for the language `#1'. Using the pattern for}%
\typeout{** the default language instead.}%
\else
\language=\csname l@#1\endcsname
\fi
#2}}

\bibitem{sajid2012multi}
Q.~Sajid, R.~Luna, and K.~Bekris, ``Multi-agent pathfinding with simultaneous
  execution of single-agent primitives,'' in \emph{International symposium on
  combinatorial search}, vol.~3, no.~1, 2012.

\bibitem{erdmann1987multiple}
M.~Erdmann and T.~Lozano-Perez, ``On multiple moving objects,''
  \emph{Algorithmica}, vol.~2, pp. 477--521, 1987.

\bibitem{wagner2015subdimensional}
G.~Wagner and H.~Choset, ``Subdimensional expansion for multirobot path
  planning,'' \emph{Artificial intelligence}, vol. 219, pp. 1--24, 2015.

\bibitem{sharon2015conflict}
G.~Sharon, R.~Stern, A.~Felner, and N.~R. Sturtevant, ``Conflict-based search
  for optimal multi-agent pathfinding,'' \emph{Artificial Intelligence}, vol.
  219, pp. 40--66, 2015.

\bibitem{sartoretti2019primal}
G.~Sartoretti, J.~Kerr, Y.~Shi, G.~Wagner, T.~S. Kumar, S.~Koenig, and
  H.~Choset, ``{PRIMAL}: {P}athfinding via {R}einforcement and {I}mitation
  {M}ulti-{A}gent {L}earning,'' \emph{IEEE Robotics and Automation Letters
  (RA-L)}, vol. 4(3), pp. 2378--2385, 2019.

\bibitem{damani2021primal}
M.~Damani, Z.~Luo, E.~Wenzel, and G.~Sartoretti, ``{PRIMAL}$_2$: {P}athfinding
  via {R}einforcement and {I}mitation {M}ulti-{A}gent {L}earning -
  {L}ifelong,'' \emph{IEEE RA-L}, vol.~6, no.~2, pp. 2666--2673, 2021.

\bibitem{aamas2023scrimp}
Y.~Wang, B.~Xiang, S.~Huan, and G.~Sartoretti, ``{SCRIMP}: Scalable
  communication for reinforcement- and imitation-learning-based multi-agent
  pathfinding,'' in \emph{\textbf{Accepted as an extended abstract} (poster
  presentation) at AAMAS}, 2023.

\bibitem{wang2022distributed}
Y.~Wang, M.~Damani, P.~Wang, Y.~Cao, and G.~Sartoretti, ``Distributed
  reinforcement learning for robot teams: A review,''
  \emph{\textit{Conditionally Accepted to} Springer's Current Robotics
  Reports}, 2022.

\bibitem{shaw2022formic}
S.~Shaw, E.~Wenzel, A.~Walker, and G.~Sartoretti, ``{ForMIC}: {For}aging via
  {M}ultiagent rl with {I}mplicit {C}ommunication,'' \emph{IEEE Robotics and
  Automation Letters}, vol.~7, no.~2, pp. 4877--4884, 2022.

\bibitem{zhang2019efficient}
S.~Q. Zhang, Q.~Zhang, and J.~Lin, ``Efficient communication in multi-agent
  reinforcement learning via variance based control,'' \emph{Advances in Neural
  Information Processing Systems}, vol.~32, 2019.

\bibitem{foerster2018counterfactual}
J.~Foerster, G.~Farquhar, T.~Afouras, N.~Nardelli, and S.~Whiteson,
  ``Counterfactual multi-agent policy gradients,'' in \emph{Proceedings of the
  AAAI conference on artificial intelligence}, vol.~32, no.~1, 2018.

\bibitem{rashid2018qmix}
T.~Rashid, M.~Samvelyan, C.~Schroeder, G.~Farquhar, J.~Foerster, and
  S.~Whiteson, ``Qmix: Monotonic value function factorisation for deep
  multi-agent reinforcement learning,'' in \emph{ICML}, 2018, pp. 4295--4304.

\bibitem{li2021message}
Q.~Li, W.~Lin, Z.~Liu, and A.~Prorok, ``Message-aware graph attention networks
  for large-scale multi-robot path planning,'' \emph{IEEE Robotics and
  Automation Letters}, vol.~6, no.~3, pp. 5533--5540, 2021.

\bibitem{ma2021distributed}
Z.~Ma, Y.~Luo, and H.~Ma, ``Distributed heuristic multi-agent path finding with
  communication,'' in \emph{ICRA}, 2021, pp. 8699--8705.

\bibitem{ma2021learning}
Z.~Ma, Y.~Luo, and J.~Pan, ``Learning selective communication for multi-agent
  path finding,'' \emph{IEEE RA-L}, vol. 7(2), pp. 1455--1462, 2021.

\bibitem{li2022multi}
W.~Li, H.~Chen, B.~Jin, W.~Tan, H.~Zha, and X.~Wang, ``Multi-agent path finding
  with prioritized communication learning,'' \emph{arXiv preprint
  arXiv:2202.03634}, 2022.

\bibitem{stern2019multi}
R.~Stern, N.~Sturtevant, A.~Felner, S.~Koenig, H.~Ma, T.~Walker, J.~Li,
  D.~Atzmon, L.~Cohen, T.~Kumar, \emph{et~al.}, ``Multi-agent pathfinding:
  Definitions, variants, and benchmarks,'' in \emph{Proceedings of SoCS},
  vol.~10, no.~1, 2019, pp. 151--158.

\bibitem{vaswani2017attention}
A.~Vaswani, N.~Shazeer, N.~Parmar, J.~Uszkoreit, L.~Jones, A.~N. Gomez,
  {\L}.~Kaiser, and I.~Polosukhin, ``Attention is all you need,''
  \emph{Advances in neural information processing systems}, vol.~30, 2017.

\bibitem{wang2022fcmnet}
Y.~Wang and G.~Sartoretti, ``{FCMNet}: {F}ull {C}ommunication {M}emory {Net}
  for team-level cooperation in multi-agent systems,'' in \emph{AAMAS}, 2022,
  pp. 1355--1363.

\bibitem{mishra2017simple}
N.~Mishra, M.~Rohaninejad, X.~Chen, and P.~Abbeel, ``A simple neural attentive
  meta-learner,'' \emph{arXiv preprint arXiv:1707.03141}, 2017.

\bibitem{parisotto2020stabilizing}
E.~Parisotto, F.~Song, J.~Rae, R.~Pascanu, C.~Gulcehre, S.~Jayakumar,
  M.~Jaderberg, R.~L. Kaufman, A.~Clark, S.~Noury, \emph{et~al.}, ``Stabilizing
  transformers for reinforcement learning,'' in \emph{International conference
  on machine learning}.\hskip 1em plus 0.5em minus 0.4em\relax PMLR, 2020, pp.
  7487--7498.

\bibitem{schulman2017proximal}
J.~Schulman, F.~Wolski, P.~Dhariwal, A.~Radford, and O.~Klimov, ``Proximal
  policy optimization algorithms,'' \emph{arXiv preprint arXiv:1707.06347},
  2017.

\bibitem{sunehag2017value}
P.~Sunehag, G.~Lever, A.~Gruslys, W.~M. Czarnecki, V.~Zambaldi, M.~Jaderberg,
  M.~Lanctot, N.~Sonnerat, J.~Z. Leibo, K.~Tuyls, \emph{et~al.},
  ``Value-decomposition networks for cooperative multi-agent learning,''
  \emph{arXiv preprint arXiv:1706.05296}, 2017.

\bibitem{li2020new}
J.~Li, G.~Gange, D.~Harabor, P.~J. Stuckey, H.~Ma, and S.~Koenig, ``New
  techniques for pairwise symmetry breaking in multi-agent path finding,'' in
  \emph{ICAPS}, vol.~30, 2020, pp. 193--201.

\bibitem{yang2021exploration}
T.~Yang, H.~Tang, C.~Bai, J.~Liu, J.~Hao, Z.~Meng, and P.~Liu, ``Exploration in
  deep reinforcement learning: a comprehensive survey,'' \emph{arXiv preprint
  arXiv:2109.06668}, 2021.

\bibitem{amin2021survey}
S.~Amin, M.~Gomrokchi, H.~Satija, H.~van Hoof, and D.~Precup, ``A survey of
  exploration methods in reinforcement learning,'' \emph{arXiv preprint
  arXiv:2109.00157}, 2021.

\bibitem{badia2020agent57}
A.~P. Badia, B.~Piot, S.~Kapturowski, P.~Sprechmann, A.~Vitvitskyi, Z.~D. Guo,
  and C.~Blundell, ``Agent57: Outperforming the atari human benchmark,'' in
  \emph{International Conference on Machine Learning}.\hskip 1em plus 0.5em
  minus 0.4em\relax PMLR, 2020, pp. 507--517.

\bibitem{savinov2018episodic}
N.~Savinov, A.~Raichuk, R.~Marinier, D.~Vincent, M.~Pollefeys, T.~Lillicrap,
  and S.~Gelly, ``Episodic curiosity through reachability,'' \emph{arXiv
  preprint arXiv:1810.02274}, 2018.

\end{thebibliography}

\clearpage

\section*{Supplementary Material}
\section{Experimental Setup}

All training instances (i.e., everything but comparison experiments) were conducted on a server equipped with four Nvidia GeForce RTX 3090 GPUs and one Intel(R) Core(TM) i9-10980XE CPU (18 cores, 36 threads).
Only one GPU is used during training.
The code for the neural network part was written in torch 1.9.0.
We relied on Ray 1.2.0 to employ 16 processes to collect data in parallel.
The training speed shows a significant improvement compared to our previous work, and the model using standard hyperparameters converges within 8 hours on tasks with 8 agents, but longer training can lead to better scalability.
Our full training and testing code is available at \url{https://github.com/marmotlab/SCRIMP}.

\section{Hyperparameters}

Table~\ref{table:hyperparameter} presents the hyperparameters used to train the model evaluated in Section V.A of our paper.
The first 20 hyperparameters are general hyperparameters for PPO and our MAPF task, while remaining hyperparameters are unique to the methods proposed in our paper.

\begin{table}[h]
\caption{Hyperparameters table}
\label{table:hyperparameter}
\scalebox{0.8}{
\begin{tabular}{l|l}
\hline
Hyperparameter & Value \\ \hline \hline
Number of agents & 8 \\ \hline
FOV size & 3 \\ \hline
Maximum episode length & 256 \\ \hline
World size & (10,40) \\ \hline
Obstacle density & (0\%,50\%) \\ \hline
Learning rate & 1e-5 \\ \hline
Discount factor & 0.95 \\ \hline
Gae lamda & 0.95 \\ \hline
Clip parameter for probability ratio $\epsilon$ & 0.2 \\ \hline
Gradient clip norm & 10 \\ \hline
Number of epoch & 10 \\ \hline
Number of processes & 16 \\ \hline
Mini batch size & 1024 \\ \hline
Imitation learning rate & 10\% \\ \hline
Optimizer & AdamOptimizer \\ \hline
Entropy coefficient & 0.01 \\ \hline
Policy coefficient & 10 \\ \hline
Valid action coefficient & 0.5 \\ \hline
Block coefficient & 0.5 \\ \hline
Number of seeds in main processing & 1234 \\ \hline
Dimension of the feed-forward network of the encoder & 1024 \\ \hline
Number of computation blocks & 1 \\ \hline
Number of heads & 8 \\ \hline
Dimension of q,k,v & 32 \\ \hline
Distance discount factor $\mu$ & 0.1 \\ \hline
Intrinsic value coefficient & 0.08 \\ \hline
Extrinsic value coefficient & 0.08 \\ \hline
Capacity of episodic buffer $M$ & 80 \\ \hline
Threshold for obtaining intrinsic reward $\tau$ & random(1,3) \\ \hline
Threshold for adding to buffer $\rho$ & 3 \\ \hline
Number of steps to start adding intrinsic rewards & 1e6 \\ \hline
$\varphi$ & 0.2 \\ \hline
$\beta$ & 1 \\ \hline
\end{tabular}}
\vspace{-0.5cm}
\end{table}

\begin{table*}[h]
\centering
\caption{Extended experimental results of Section V.A}
\label{table:additional performance}
\scalebox{0.65}{
\begin{tabular}{|l|lll|lll|lll|lll|}
\hline
\multicolumn{1}{|c|}{\multirow{2}{*}{Methods}} & \multicolumn{12}{c|}{16 agents in $20\times20$ world with 0\%, 15\%, 30\% static obstacle rate} \\ \cline{2-13} 
\multicolumn{1}{|c|}{} & \multicolumn{3}{c|}{EL ↓} & \multicolumn{3}{c|}{MR ↑} & \multicolumn{3}{c|}{CO ↓} & \multicolumn{3}{c|}{SR ↑} \\ \hline
ODrM* &27.38(3.63) & 27.64(3.74) & 34.61(6.55) & - & - & - & 0.00\%(0.00\%) & 0.00\%(0.00\%) & 0.00\%(0.00\%) &  100\% &  100\%& 95\% \\
DHC &29.15(4.66)& 34.72(8.44)&55.71(26.75) & \textbf{16.00(0.00)} & \textbf{16.00(0.00)} & 15.80(0.49) & \textbf{0.00\%(0.00\%)} &\textbf{0.00\%(0.00\%)} &\textbf{0.00\%(0.00\%) }&\textbf{100\%}& \textbf{100\%} &83\%   \\
PICO &29.83(5.80)& 61.83(30.73)& -& \textbf{16.00(0.00)} &13.85(1.92)  & 9.96(2.17) & \textbf{0.00\%(0.00\%)} &\textbf{0.00\%(0.00\%)} &\textbf{0.00\%(0.00\%)} &\textbf{100\%}&18\%  & 0\% \\
SCRIMP &\textbf{26.40(3.87)}&\textbf{27.75(4.34)} &\textbf{36.00(8.75)} & \textbf{16.00(0.00)} & \textbf{16.00(0.00)} &\textbf{16.00(16.00)}  & \textbf{0.00\%(0.00\%)} &0.02\%(0.17\%)&0.1\%(0.21\%) &\textbf{100\%} &\textbf{100\%}& \textbf{100\%}  \\\hline

 & \multicolumn{12}{c|}{64 agents in $40\times40$ world with 0\%, 15\%, 30\% static obstacle rate} \\ \hline
 ODrM* & 61.40(5.33) & 62.42(6.11) & 73.59(10.25)  & - & - & - & 0.00\%(0.00\%) & 0.00\%(0.00\%)  & 0.00\%(0.00\%) & 100\% & 100\%  & 92\% \\
DHC &65.59(7.22)&82.6(22.11)& 137.87(39.15)& \textbf{64.00(0.00)} & \textbf{63.99(0.10)} & 61.44(4.42) &\textbf{0.00\%(0.00\%)}  &\textbf{0.00\%(0.00\%)} &\textbf{0.00\%(0.00\%)} &\textbf{100\%}&\textbf{99\%} & 46\% \\
PICO &67.67(9.97)&- &- &61.64(1.29)  &47.52(3.60)  & 27.61(5.23) &\textbf{0.00\%(0.00\%)}  &0.003\%(0.004\%) &0.07\%(0.05\%) &6\%& 0\% &0\% \\
SCRIMP & \textbf{61.82(6.13)}&\textbf{65.44(10.28)} &\textbf{90.72(29.16)} &\textbf{64.00(0.00)}  &\textbf{63.99(0.10)}  &\textbf{63.77(1.23)} & \textbf{0.00\%(0.00\%)} & 0.02\%(0.03\%)&0.56\%(1.17\%) &\textbf{100\%}& \textbf{99\% }&\textbf{89}\%\\\hline

 & \multicolumn{12}{c|}{256 agents in $40\times40$ world with 0\%, 15\%, 30\% static obstacle rate} \\ \hline
 ODrM* & 69.77(5.77) & 76.08(6.57) & -  & - & - & - & 0.00\%(0.00\%) & 0.00\%(0.00\%)  & 0.00\%(0.00\%) & 96\% & 39\%  & 0\% \\
SCRIMP & 77.26(7.03)&104.87(24.21) &243.5(5.5) &256.00(0.00) &255.95(0.22) &181.30(43.89) & 0.07\%(0.03\%) & 0.85\%(0.52\%)&16.54\%(7.49\%) &100\%& 95\% &2\%\\\hline

\end{tabular}}
\vspace{-0.1cm}
\end{table*}

\section{Extended experiments}

Using the same environment and hyperparameters described in Section V.A of our paper, we further tested our learned models in additional experiments with different team sizes, world sizes and obstacle densities as shown in Table~\ref{table:additional performance}.
The results of these extended experiments are aligned with the experimental analysis in Section V.A of our paper.

\section{Influence of FOV size}

\begin{figure}[h]
  \centering
  \setlength{\belowcaptionskip}{-0.3cm}
  \includegraphics[width=0.5\columnwidth]{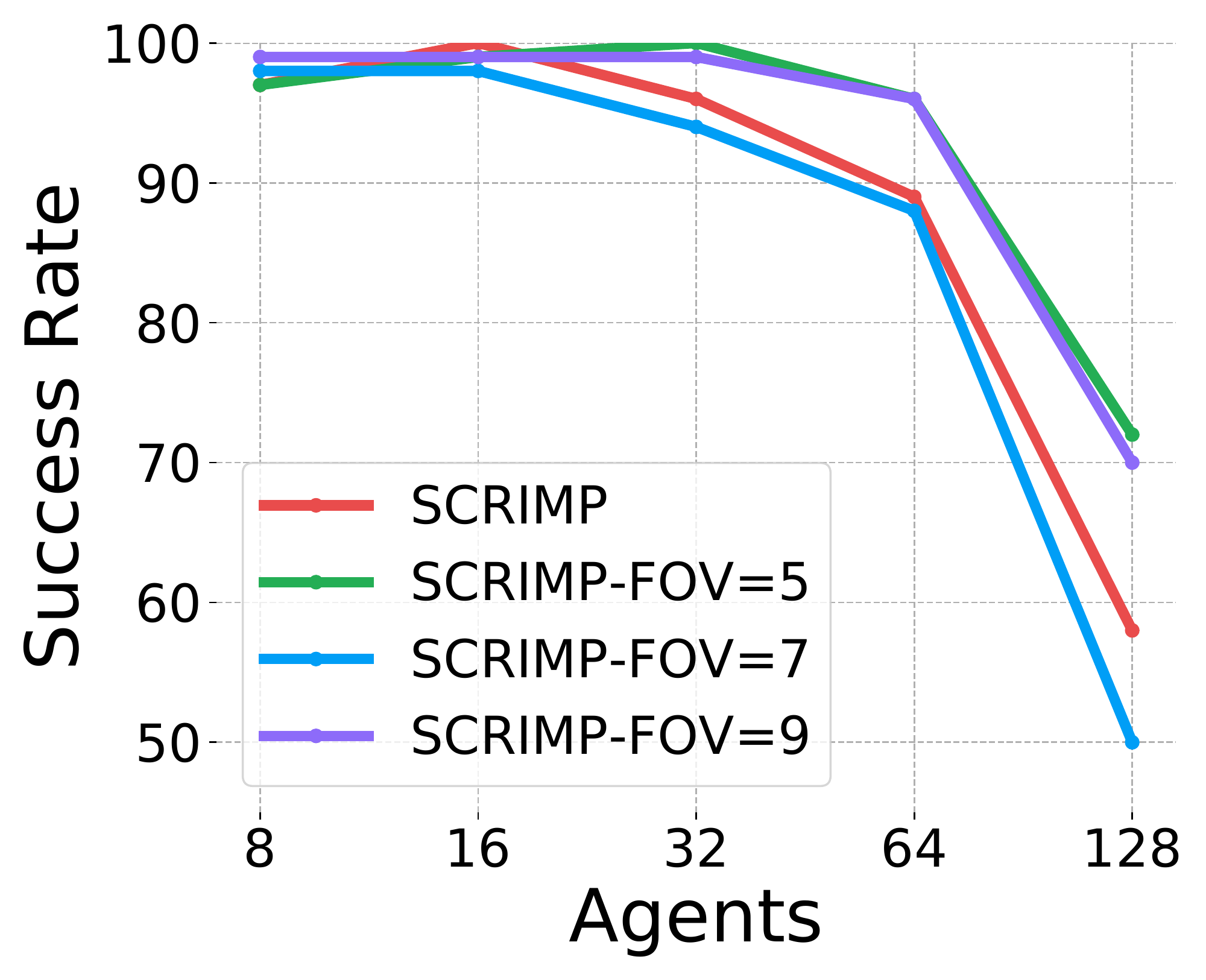}
  \caption{Success rate of SCRIMP variants with different FOV dimensions.
  The size of the world changes with the number of agents as described in the paper, but obstacle density remains 30\%.}
 \label{fig:fov}
  \vspace{-0.3cm}
\end{figure}

We further tested the performance of our model with different FOV sizes.
Figure~\ref{fig:fov} shows that in tasks involving 16 or fewer agents, the benefit of expanding the FOV is not significant, as the communication mechanism of SCRIMP already provides sufficient information.
The best performance is achieved when the FOV size is 5, where we see a 72\% success rate in the 128-agent task, in a $40\times40$ world with 30\% obstacles.
As expected, the time required for convergence increases with the FOV size, since a larger FOV requires a larger neural network with more trainable weights.

\end{document}